\documentclass[10pt,twocolumn,letterpaper]{article}

\usepackage{cvpr}              

\usepackage{graphicx}
\usepackage{amsmath}
\usepackage{amssymb}
\usepackage{booktabs}
\usepackage[T1]{fontenc}
\usepackage{mathtools}

\usepackage[pagebackref,breaklinks,colorlinks]{hyperref}

\usepackage[capitalize]{cleveref}
\crefname{section}{Sec.}{Secs.}
\Crefname{section}{Section}{Sections}
\Crefname{table}{Table}{Tables}
\crefname{table}{Tab.}{Tabs.}

\def\cvprPaperID{3864} 
\def\confName{CVPR}
\def\confYear{2023}

\begin{document}
	
	\title{Lossy Compression for Robust Unsupervised Time-Series Anomaly Detection}
	
	\author{Christopher P. Ley\\
		Advanced Center for Electrical and Electronic Engineering\\
		General Bari 699, Valparaíso, Chile\\
		{\tt\small info@christopherley.com}
	\and
	Jorge F. Silva\\
	University of Chile\\
	Beauchef 850, Santiago, Chile\\
	{\tt\small josilva@ing.uchile.cl}
}
\maketitle

\begin{abstract}
	A new Lossy Causal Temporal Convolutional Neural Network Autoencoder for anomaly detection is proposed in this work.  Our framework uses a rate-distortion loss and an entropy bottleneck to learn a compressed latent representation for the task. The main idea of using a rate-distortion loss is to introduce representation flexibility that ignores or becomes robust to unlikely events with distinctive patterns, such as anomalies.  These anomalies manifest as unique distortion features that can be accurately detected in testing conditions. This new architecture allows us to train a fully unsupervised model that has high accuracy in detecting anomalies from a distortion score despite being trained with some portion of unlabelled anomalous data. This setting is in stark contrast to many of the state-of-the-art unsupervised methodologies that require the model to be only trained on "normal data". We argue that this partially violates the concept of unsupervised training for anomaly detection as the model uses an informed decision that selects what is normal from abnormal for training. Additionally, there is evidence to suggest it also effects the models ability at generalisation. We demonstrate that models that succeed in the paradigm where they are only trained on normal data fail to be robust when anomalous data is injected into the training. In contrast, our compression-based approach converges to a robust representation that tolerates some anomalous distortion. The robust representation achieved by a model using a rate-distortion loss can be used in a more realistic unsupervised anomaly detection scheme.
\end{abstract}

\section{Introduction}
\label{sec:intro}
\textit{Anomalous data} can be a dubious concept to define, it is often a subjective term, highly domain specific and often defined with handcrafted features. More commonly, an anomaly is easier to define as the antithesis of \textit{normal}.

Weakly supervised and unsupervised approaches have been proposed to solve anomaly detection problems\cite{akcay2018ganomaly,chen2022utrad,geiger2020tadgan,le2020learning,liu2022time,macikag2021unsupervised,zhang2021soft}. Unsupervised approaches to anomaly detection, much of those that have attracted research attention\cite{akcay2018ganomaly, macikag2021unsupervised, geiger2020tadgan}, aim to detect anomalies using prior knowledge only from normal samples. 

Most modern detection methods are mainly based on reconstruction models like Autoencoders (AE)\cite{baur2018deep,bergmann2018improving, solch2016variational} and Generative Adversarial Networks (GAN) \cite{akcay2018ganomaly,geiger2020tadgan,liu2022time,schlegl2019f,schlegl2017unsupervised}. These approaches train the reconstruction model with only normal samples based on the assumption of generalisation gap, which means the reconstruction succeeds with only normal samples but fails with anomalies. However, such a prior forces reconstruction-based methods to suffer from several problems. Firstly, it is hard to learn the expressive representations when the training data are only normal samples without diversity\cite{salehi2021multiresolution}, this brings training instability problems. Second, there is no guarantee that the generality gap exists in these reconstruction-based models\cite{chen2022utrad} that are trained only on a normal prior. Under-trained AEs may not be able to reconstruct the normal signal while a over-trained AE may even reconstruct the abnormal signals. A truly unsupervised model, we argue, should be robust to unlabelled anomalies and have the capability of excluding them from training data and still be effective at detecting them during reconstruction. A strategy that doesn't require an explicit normal prior has large practical relevance as well, it has been noted in the past\cite{Kingma2014} that typically an extreme lack of anomalies causes a severe class imbalance in any labelled approaches. Consequently, an algorithm with the capacity to be robust to unlabelled anomalies could relevant for a large class of industrial anomaly detection problems where a significant section of the data is unlabelled.

In this work, we claim that lossy compression (in the classical rate-distortion sense) can be used as a learning approach for unsupervised anomaly detection by permitting training on distorted (anomalous) data. The main contribution of this work is demonstrating the role that an entropy constraint (an entropy bottleneck)\cite{balle2018variational,Dubois2021,kato2020rate} plays in learning robust compressed features of normal time-series signals in an unsupervised way. Our strategy offers a methodology to design a Variational Auto Encoder\cite{Alemi2017,Kingma2014,Tishby2015} that, for a given rate constraint (in bits per sample), allows expressive representations of time-series patterns (within a predefined slight distortion). These representations have the added capacity to detect abnormal behaviours (in the form of complex outlier features on the distortion) with high accuracy. We conjecture that normal statistical patterns are redundant in many practical domains, particularly for high-dimensional time-series data. Exploiting this redundancy in the classical rate-distortion sense offers the possibility of designing an encoder-decoder that captures what is essential to reconstruct \textit{normal} statistical behaviour.

The compressed optimality of the representations (attributed to the role played by the entropy bottleneck in the optimisation)\cite{Achille2018,silva2022interplay,Tishby2015} makes them insensitive (invariant) to abnormal features non-seen systematically during training. In that regard, the framework learns optimal encoder-decoders of normal behaviours that, at the same time, are blind to anomalies, these abnormal features can then be sharply detected using the distortion after reconstruction. Furthermore, our experimental analysis shows that this entropy-driven invariant property makes our solution robust to the presence of outliers (abnormal examples) in training. This means that our encoder-decoders can be trained unsupervised, i.e., we could tolerate a small proportion of non-label abnormal examples during training.

On the learning task, we follow the variational approach utilising an entropy bottleneck proposed by Balle \etal \cite{balle2018variational}, with the decomposed rate-distortion loss of Kato \etal \cite{kato2020rate}, to learn an encoder-decoder that operates optimally in the rate-distortion plane. For this task, we propose using a temporal causal convolution \cite{oord2016wavenet,thill2020time} architecture for the encoder and decoder aligned to the type of prior causal sequential structure observed in complex time-series data. Finally, we propose a strategy based on detecting abnormal distortion trends for the detection by comparing the original signal with its lossy reconstruction obtained after decompression.  Our results show that we could achieve state-of-the-art performance unsupervised in anomaly detection on the Skoltech Anomaly Benchmark (SKAB)\cite{skab} even after injecting 5\% anomalous data into training set (unlabelled).

The rest of the paper is organized as follows:
\begin{itemize}
	\item In \Cref{sec:lossy} we highlight how lossy compression can be used with variational models in the context of anomaly detection and outline how the \emph{rate-distortion optimisation} (RDO) objective is used to achieve a compressed representation.
	\item In \Cref{sec:TCN} we elaborate on the Temporal Convolutional Neural Network Autoencoder (TCN-AE) model used in this paper and specifically how the causal dilated convolution operations and \emph{entropy bottleneck} are incorporated into the Autoencoder architecture.
	\item In \Cref{sec:experiments} we detail the experimental procedure, with \Cref{sec:data} outlining the data-set used for experimental validation, \Cref{sec:practical_anomaly_detection} detailing how a TCN-AE can be used in the context of anomaly detection on time-series data for one sample (1-shot), \Cref{sec:1-shot_results} presents the 1-shot detection results and \Cref{sec:multi-shot} details how the anomaly detection can be expanded to detection over a whole time-series under a multi-shot detection scheme.
	\item Finally in \Cref{sec:analysis} we discuss our results\footnote{Code is available at https://}
\end{itemize}

\section{Lossy Compression with Variational Models for Anomaly Detection}\label{sec:lossy}

In transform-based lossy compression \cite{Gray1990,balle2018variational}, the encoder transforms the input signal vector $\mathbf{x}$ using a parametric transform $f_{\boldsymbol\theta}(\mathbf{x})$ into a latent representation $\mathbf{y}$, which is then quantised to form $\mathbf{\hat{y}}$. Because $\mathbf{\hat{y}}$ is discrete-valued (given the rate constrain), it can be losslessly compressed using entropy coding\cite{Cover2006} and transmitted as a sequence of bits. On the other side, the decoder recovers $\mathbf{\hat{y}}$ from the compressed signal, and subject to a parametric transform $g_{\boldsymbol\phi}(\mathbf{\hat{y}})$, recovers the reconstructed signal vector $\mathbf{\hat{x}}$. In practice, the parametric transforms $f_{\boldsymbol\theta}$ and $g_{\boldsymbol\phi}$ are expressive parameterised functions. In our context,  we are referring to artificial neural networks (ANN) and the parameters $\boldsymbol\theta$ and $\boldsymbol\phi$ encapsulate the weights of the neurons \etc (see \Cref{sec:TCN} for details)

The vector quantisation (VQ) induces errors in the reconstruction, which is tolerated in the context of lossy compression, giving rise to the celebrated \emph{rate-distortion optimisation} problem \cite{Cover2006,Gersho1992,Gray1990}.  Following the clever formalisation proposed by Ball{\'e} \etal \cite{balle2018variational}, the rate is represented by the expected code length (in bits per sample) of the compressed representation that can be interpreted as a cross-entropy:
\begin{equation}\label{eq:rate}
	R = \mathbb{E}_{\mathbf{x}\sim p_{\mathbf{x}}}\left[-\log_2p_{\mathbf{\hat{y}}}\left(Q(g_{\boldsymbol\phi}(\mathbf{x}))\right)\right].
\end{equation}
In this expression, $Q(\cdot)$ represents the VQ function, and $p_{\mathbf{\hat{y}}}$ is the entropy model (a probability). In its original implementation, Ball{\'e} \etal \cite{balle2018variational} defines the distortion $D$ as the expected difference between the reconstruction $\mathbf{\hat{x}}$ and the original signal $\mathbf{x}$, as measured by some distortion metric $\rho(\cdot)$ \eg a norm (or squared error). 
\begin{equation}\label{eq:balle_distortion_loss}
	D(\mathbf{x},\mathbf{\hat{x}}) = \mathbb{E}_{\mathbf{x}\sim p_{\mathbf{x}}}\rho(\mathbf{x},\mathbf{\hat{x}})
\end{equation}
The coarseness of the quantisation, or alternatively, the warping of the representation implied by both the encoder and decoder transforms, affects both rate and distortion.  This leads to a rate-distortion trade-off, where a higher rate allows for lower distortion, and visa versa. Formally \cite{balle2018variational} parameterise the problem by $\lambda$, a weight on the distortion term, leading to the rate-distortion loss function:
\begin{equation}
	L = R + \lambda D
\end{equation}
Kato \etal \cite{kato2020rate} presented a further decomposition of the distortion measure.  Based on the work of Rolinek \etal \cite{rolinek2019variational}, they highlight the equivalency with the $\beta$-VAE (Variational Autoencoder \cite{chen2018isolating,higgins2016beta,kim2018disentangling}) where weight $\lambda_1$ controls the degree of reconstruction, and $\lambda_2$ ($\approxeq \beta^{-1}$ of $\beta$-VAE) controls a scaling between data and latent spaces respectively such that 
\begin{equation}
	\lambda D(\mathbf{x}, \mathbf{\hat{x}}) \approxeq \lambda_1 D(\mathbf{x}, \mathbf{\tilde{x}}) + \lambda_2 D(\mathbf{\hat{x}}, \mathbf{\tilde{x}}).
\end{equation}
Here $\mathbf{\tilde{x}}$ is the reconstructed of the signal $\mathbf{x}$ from the latent representation $\mathbf{y}$ using the decoder $g_{\boldsymbol\phi}(\mathbf{y})$ (identical parameterisation) without quantisation. Thus $D(\mathbf{\hat{x}}, \mathbf{\tilde{x}})$ is a measure of the distortion induced simply due to quantisation and $D(\mathbf{x}, \mathbf{\tilde{x}})$ is the reconstruction error. The \emph{rate-distortion optimisation} loss is then defined as follows
\begin{equation}\label{eq:rdo_loss}
	L = -\log\left(P_{\mathbf{z}, \boldsymbol\psi}(\mathbf{z})\right) + \lambda_1D(\mathbf{x}, \mathbf{\hat{x}}) + \lambda_2D(\mathbf{\hat{x}}, \mathbf{\tilde{x}})
\end{equation}
where $-\log\left(P_{\mathbf{z}, \boldsymbol\psi}(\mathbf{z})\right)$ is the estimated rate of the latent variable (the compression term). In this work, we follow the \emph{entropy bottleneck} developed in Ball{\'e} \etal \cite{balle2018variational} to estimate the optimum rate through the use a univariate independent (factorised) model $P_{\mathbf{z}, \boldsymbol\psi}(\mathbf{z}) = \prod_{i=1}^{N}P_{z_i, \boldsymbol\psi}(z_i)$ and a soft (differentiable) uniform quantisation for the latent variable (see \cite{balle2018variational} for more details). \Cref{fig:tcn} demonstrates how the \emph{entropy bottleneck} is integrated into the autoencoder. 
\subsection{Anomaly Detection}\label{sec:anomaly}
Given a information source (or input) $\mathbf{x}$, containing a large number of normal samples, we can formalise the problem of detecting anomalies as an binary hypothesis testing problem where:
\begin{itemize}
	\item $H_0:$  $\mathbf{x}\sim \mu_x$ (normal statistical behaviour),
	\item $H_1:$ $\mathbf{x}\sim q_x \neq \mu_x$ (abnormal). 
\end{itemize}
The probability $\mu_{\mathbf{x}}$ models the normal statistical pattern of $\mathbf{x}$ where the condition in $H_1$ represents a deviation from $\mu_{\mathbf{x}}$ represented by any probability $q_x$ that is different than $\mu_x$.  

Our framework uses the encoder and decoder obtained as the solution of the rate-distortion loss optimisation in \cref{eq:rdo_loss} in the training stage, while anomaly detection is performed during the inference (testing) stage using the reconstruction error of the observed signal. More precisely, a signal $\mathbf{x}$ is fed through the Temporal Convolutional Neural Network Autoencoder (TCN-AE) (see \Cref{sec:TCN} for details) and an estimated signal $\mathbf{\hat{x}}$ is reconstructed and determined to be anomalous if $|\mathbf{x} - \mathbf{\hat{x}}| > \delta$, where $\delta$ is the determined threshold for an anomalous signal. In practice this is much more nuanced, and is elaborated in \Cref{sec:practical_anomaly_detection}, as the reconstructed signal under the TCN-AE is a large sliding window ($T=200$ samples) with a multiple channel output (8 in our case).
\section{Temporal Convolutional Neural Network Autoencoder}\label{sec:TCN}
Here we present the specific design choices of the encoder and decoder used in the optimization of the rate-distortion loss $L$ in  \cref{eq:rdo_loss}. 
\begin{figure*}
	\centering
	\includegraphics[width=1.0\linewidth]{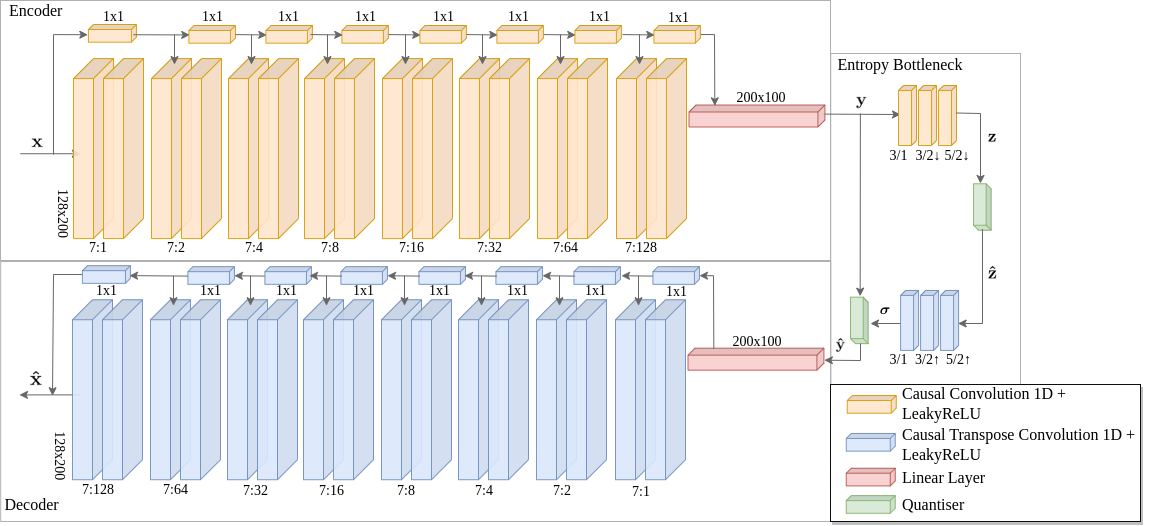}
	\caption{Architecture of the Lossy Causal Temporal Convolutional Neural Network Autoencoder. The Encoder and Decoder are similar architectures both implement causal temporal stacked 1D convolutions (transposed convolution for decoder), see \Cref{fig:dialatedcovolutions}. Each layer spans over an input length (in time) of 200 sample with 128 channels per layer. The legend below each convolutional layer is in the form $kernel\; width : dialation\;size$. The \emph{entropy bottleneck} is identical to that presented in Ball{\'e} \cite{balle2018variational} below each convolutional (and transposed convolutional) layer is a key of the form $kernel\; width / stride$ with a down arrow $\downarrow$ indicating down-sampling and an up arrow $\uparrow$ signifying up-sampling conversely. The vector $\mathbf{z}$ is result of the parametric transform and $\boldsymbol\sigma$ is the spatial distributions of the standard deviations used to compress the latent variable $\mathbf{y}$.} 
	\label{fig:tcn}
\end{figure*}
We adopt the Temporal Convolutional Neural Network Autoencoder (TCN-AE).  TCN-AE is an autoencoder architecture which consists of an encoder constructed from a series of stacked convolutional layers designed to meet the following causal factorisation of a joint distribution
\cite{van2016conditional,oord2016wavenet,van2016pixel}
\begin{equation}\label{eq:conditional_prob}
	p(\mathbf{x}) = \prod_{t=1}^{T}p(x_t|x_1, \ldots, x_{t-1}).
\end{equation}
where $t$ is the time index of the vector $\mathbf{x} = \left[x_1,\ldots,x_T\right]$, in practice (such as our case) $\mathbf{x}$ can be a matrix where each entry $x_t$ is a vector of input channels such that $x_t \in \mathbb{R}^C$ and $C$ is the number of channels this would result in $\mathbf{x}\in \mathbb{R}^{C\times T}$.

\begin{figure}
	\centering
	\includegraphics[width=1.0\linewidth]{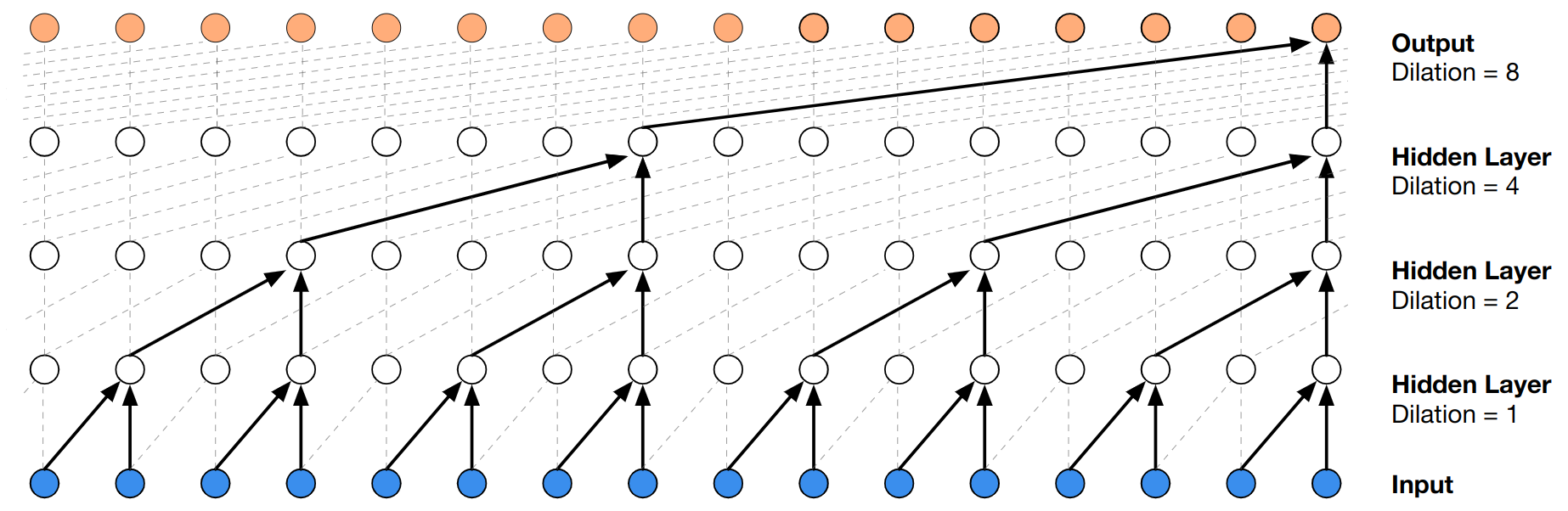}
	\caption{Visulisation of a stack of \emph{dilated} causal convolutional layers (figure from \cite{oord2016wavenet}), each consecutive layer has a $2^{\text{dilation}}$ receptive field. Notice causality is respected \ie the input on the right (the latest) only depends on those inputs in the past.}
	\label{fig:dialatedcovolutions}
\end{figure}

TCN-AE is implemented with no pooling layers and all layers are grouped into a \emph{dilated causal convolutional} introduced by Van de Oord \etal in \cite{oord2016wavenet}. Dilated causal convolutions, as displayed graphically in \Cref{fig:dialatedcovolutions}, models the causal sequential (temporal in our case) relationship between the each input per channel. This inductive prior \cite{bronstein2021geometric} (which can be thought of as a 1 dimensional directional graph or a 1D grid), allows the model to exploit the translation equivariance\cite{bronstein2021geometric} in the input signal $\mathbf{x}$ presented in  \Cref{eq:conditional_prob}. The causal convolutions ensure the model can not violate the ordering in which we present the data to the model, \ie $x_t$ can not depend on what happened in the future $x_{t+1}$ only what has occurred in the past $x_{t-1}, \ldots, x_{t-N}$. In our particular case, of time-series modelling, this resonates intuitively as at a macro scale the dynamics of a system, that generate the time-series signals, cannot violate causality \ie dynamics of the system can only be influenced by what has occurred up to time $t$ not by any future action. The dilation is a type of convolution filter where the kernel is applied over an area larger than its length by skipping input values with a certain step \cite{oord2016wavenet}\footnote{For example,  a 1D kernel (of width 3) with a dilation of 1 would convolve a weight vector $[w_1, w_2, w_3]$ over a signal, the same kernel with a dilation of 2 would apply the same operation but with a weight vector $[w_1, 0, w_2, 0, w_3]$, where the quantity of 0's separating weights is $dilation - 1$.}.

As shown in \Cref{fig:dialatedcovolutions} if we couple causal convolutions with a series of exponentially increasing dilation, this results in an exponential receptive field that grows with depth\cite{yu2015multi} allowing us to model efficiently long range dependencies in temporal data.

Finally, the TCN-AE consists of an encoder that implements an 8 block causal convolution with dilations that grow exponentially per block (of 2 layers) following $\text{dilation} = 2^{l}$ where $l\in  \left\{0,..,7\right\}$. This is illustrated in \Cref{fig:tcn}.  The decoder is architecturally identical to the encoder with the causal convolutions being replaced with causal transpose convolutions. Both encoder and decoder have residual connections per block via a $1\times1$ convolution operation. The encoder is compressed via a linear layer to the latent variable $\mathbf{y}$ which is then fed to \emph{entropy bottleneck} (see \Cref{sec:lossy} for details) which quantises the latent variable $\mathbf{\hat{y}}$ and simultaneously estimates the rate $R$ in \cref{eq:rate}. 
\section{Experiments}\label{sec:experiments}
\subsection{Dataset}\label{sec:data}
\begin{figure}
	\centering
	\includegraphics[width=1.0\linewidth]{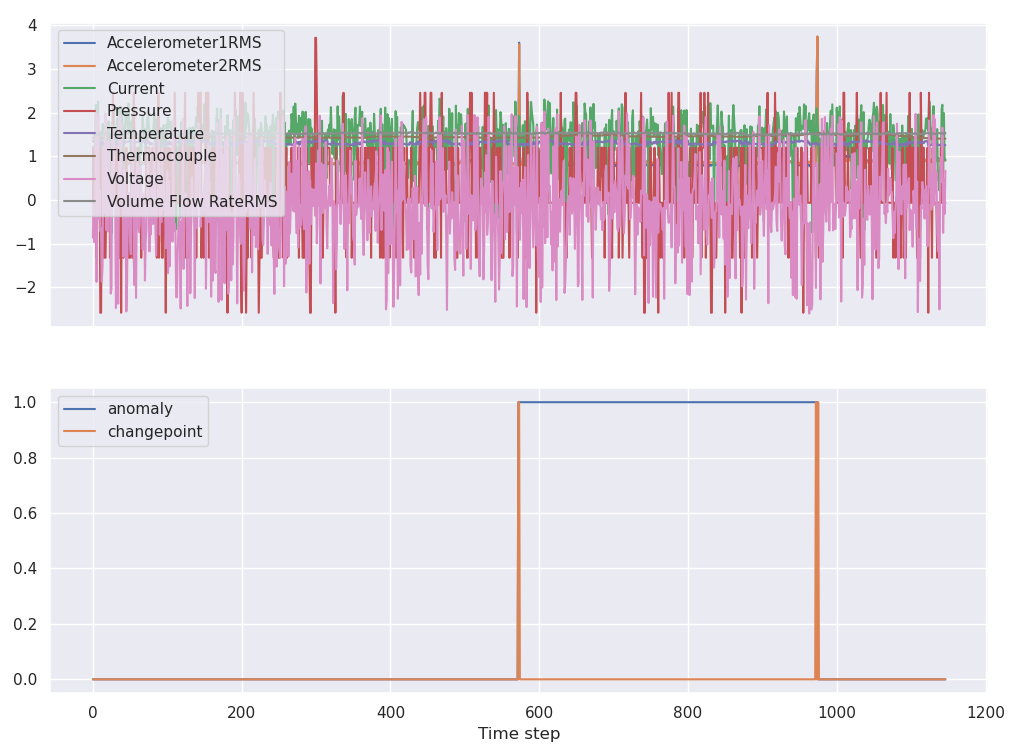}
	\caption{\emph{Skoltech Anomaly Benchmark (SKAB)}\cite{skab} set example. (1 of 34, normalised per channel)}
	\label{fig:skabdataset}
\end{figure}

All experiments were trained and tested on the \emph{Skoltech Anomaly Benchmark (SKAB)}\cite{skab} a 8 channel time-series anomaly detection benchmark, an example of which is in \Cref{fig:skabdataset}. The database contains 34 time-series sets, 1 anomaly free and the remainder beginning with normal data then after some time an anomaly is induced in the pumping system. The task is to detect the anomaly which can come from multiple sources (valves, pumps etc. see \cite{skab} for more details). We can see from \Cref{fig:skabdataset} that anomalies are not obvious.  The dataset comes from measurements of 1 single system in various points and forms (\ie different sensors) where the channels are highly correlated. Anomalous data (in the 33 anomalous sets) make up $\geq 25\%$ of each data set.

The typically scenario proposed by the benchmark itself is to treat the first section of each set (normal data) as training data and the final portion (largely anomalous) of the dataset is used for validating the performance. In order to demonstrate the validity of our hypothesis, being that we can robustly train ``unsupervised'' on completely unlabelled data, we devised a strategy where we consider 5 of the completes sets (containing anomalies) as our validation set and train our model on the remaining 26 for the whole ``normal'' portion of each set plus some percentage of the training set that is anomalous. Anomalous training portions we use ranging from $0\%$ to $25\%$ maximum. The training set is completely unlabelled such that the network is completely unaware of which elements of the data are  anomalous. 

\subsection{Practical Anomaly detection}\label{sec:practical_anomaly_detection}
We generally outlined in \Cref{sec:anomaly} that  anomaly detection is based on time-series reconstruction, and in particular it is based on the task of determining whether or not a signal is anomalous $H_1$ using the condition: 
\begin{equation}\label{eq:anomaly_detection_basic}
	d(\mathbf{x}) \coloneqq D(\mathbf{x},\mathbf{\hat{x}})= |\mathbf{x} - \mathbf{\hat{x}}| > \delta
\end{equation}
Practically in time-series signals, $\mathbf{x}$ (the original signal) consists of an interval in time $T=200$ and 8 time-series channels $C$ such that $\mathbf{x} \in \mathbb{R}^{C\times T}$. Then, $\mathbf{\hat{x}}$ is the reconstructed signal from the TCN-AE (see \Cref{sec:TCN}) such that $\mathbf{\hat{x}} \in \mathbb{R}^{C\times T}$, and $\delta$ is the anomaly threshold (detailed below). 

Both $\mathbf{x}$, $\mathbf{\hat{x}}$ matrices and as such are composed of multiple entries, where $x_{ij}$ is the $i$th row and $j$th column of $\mathbf{x}$ given $\{i,j \in \mathbb{N} :\; 0 < i \leq C\;,\;0 < j \leq T\}$
\begin{equation}
	\mathbf{x} = \begin{bmatrix}
		x_{1,1} & \cdots & x_{1,T} \\
		\vdots  & \ddots & \vdots  \\
		x_{C,1} & \cdots & x_{C,T}
	\end{bmatrix}
\end{equation}
and similarly for $\mathbf{\hat{x}}$, which is a reconstruction of $\mathbf{x}$.

During training the network learns a channel-wise normalising constant $\boldsymbol{\omega} \in \mathbb{R}^{C}$ based on the scaling the classical distortion measure $D(\mathbf{x},\mathbf{\hat{x}})$ (\cref{eq:balle_distortion_loss}) to be approximately $\mathcal{N}(0, 1)$. During inference, we scale the absolute error by $\boldsymbol{\omega}$, of which we computed from the scaled absolute error in the reconstruction (\textbf{\ae}) such that
\begin{equation}
	\text{\ae}_j = \begin{bmatrix}
		\omega_1|x_{1,j} - \hat{x}_{1,j}| \\
		\vdots  \\
		\omega_C|x_{C, j} - \hat{x}_{C, j}|
	\end{bmatrix}
\end{equation} 
$\forall j$ of $\text{\ae}_j$ we calculate the maximum which we refer to as the maximum absolute error (m\ae), then
\begin{equation}\label{eq:anomaly_detection_complete}
	\text{m\ae}_j = \max_{j}\left(\text{\ae}_j\right).
\end{equation}
where the resulting \textbf{m\ae} is a vector, such that
\begin{equation}
	\textbf{m\ae} = \left[\text{m\ae}_1, \ldots, \text{m\ae}_T\right]
\end{equation}
As we desire not only to classify the existence of an anomaly but when it occurred, we finally separate the \textbf{m\ae} vector into subsets of 10, for each \textbf{m\ae} there is $T/10=20$ subsets. Each subset is denoted $M_k$ where $\{k \in \mathbb{N} : 0 \leq k < 20\}$ then for each $k$
\begin{align}
	M_k &= \left[\text{m\ae}_{10k}, \ldots, \text{m\ae}_{10(k+1)}\right]\\
	\overline{M}_{k} &= \frac{1}{10}\sum_{n=10k}^{10(k+1)}\text{m\ae}_{n}\\
\end{align}
The the 1-shot anomaly prediction of $H_1$ becomes
\begin{equation}\label{eq:1-shot}
	d(x_k) = \begin{cases}
		1       & \quad \text{if } \overline{M}_{k} > \delta \\
		0  & \quad \text{if } \overline{M}_k \leq \delta
	\end{cases}
\end{equation}
where $x_k$ denotes detection for the interval $k$ of the input signal $\mathbf{x}$, where the $k$th interval spans $10k < j < 10(k-1), \forall i$ where $d(\mathbf{x}) = \left[d(x_0), \ldots,d(x_{19})\right]$. Typically $\delta = 1.0$ but is determined experimentally.

\subsection{1-shot Detection Results}\label{sec:1-shot_results}
\begin{figure}
	\centering
	\includegraphics[width=1.0\linewidth]{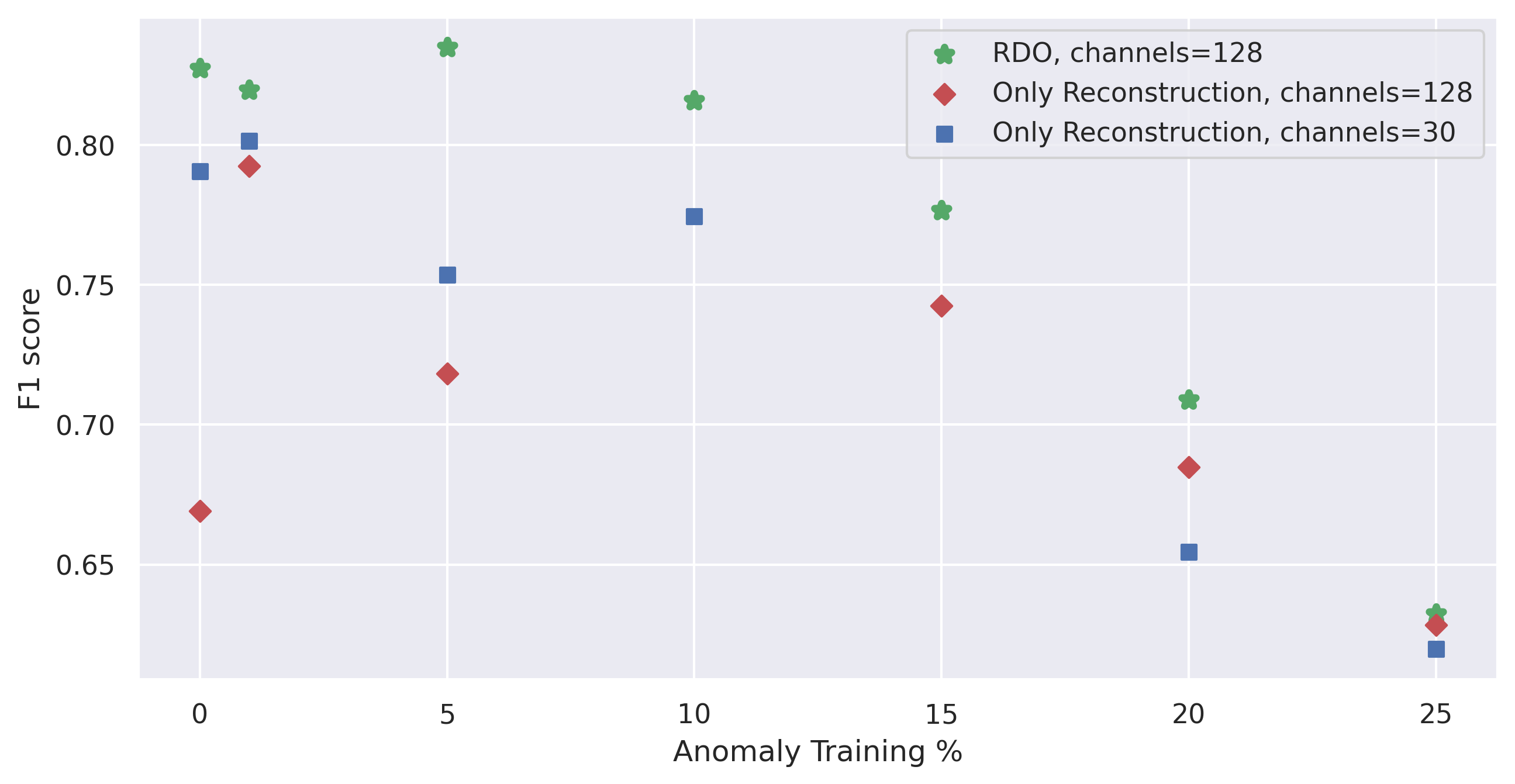}
	\caption{F1 scores for the Lossy Rate Distortion optimised (denoted RDO) \vs model trained with only distortion (mean square loss, without the entropy bottleneck). Top scoring model is trained (unsupervised) with 5\% anomalies, with $F_1$ = 0.834906 (larger is better)}
	\label{fig:f1scores}
\end{figure}

All versions of the networks used in the experiments utilise the same architecture, depicted in \Cref{fig:tcn} and detailed in \Cref{sec:TCN}. For comparison,  we trained a first baseline tests  using a standard autoencoder (AE) scheme, which we designate by ``only reconstruction'' in \Cref{fig:f1scores}. Training the AE model consists of disregarding the entropy bottleneck and feeding the latent variable $\mathbf{y}$ directly as the input of the decoder. The model is then trained by simply minimising the mean square error loss of the reconstruction $\mathbf{\hat{x}} = g_{\boldsymbol\phi}(\mathbf{y})$, i.e., the loss is equivalent to minimising the reconstruction term in \cref{eq:balle_distortion_loss} with no quantisation:
\begin{equation}
	L = D(\mathbf{x}, g_{\boldsymbol\phi}(f_{\boldsymbol\theta}(\mathbf{x})))
\end{equation}
The standard AE model was tested with the full architecture depicted in \Cref{fig:tcn} (\ie with 128 channels per layer) as well as with a reduced version with 30 channels per layer, ceteris paribus (all else being equal). Each of these models had their respective best performing $F_1$ score calculated based on the anomaly detection scheme outlined in \Cref{sec:practical_anomaly_detection} , where the $F_1$ score is given by
\begin{equation}
	F_1 = \frac{2TP}{2TP + FP + FN}.
\end{equation}\label{eq:F1} 
The results can be seen (red diamonds, and blue squares) in \Cref{fig:f1scores}. 

Concerning our rate-distortion optimised (RDO) models,  these were trained using 
the loss in \cref{eq:rdo_loss} including the \emph{entropy bottleneck}.  The results are depicted with green stars in \Cref{fig:f1scores}.
\Cref{tab:results} shows that the $F_1$ scores for the lossy model (RDO) are relatively stable (robust) when adding up to $10\%$ anomalous data (unlabelled) to our training set.  In contrast, both AE model's anomaly detection performance degrade significantly as the training data is corrupted with anomalous examples. Remarkably, the RDO model outperforms all other models in absolute anomaly detection ($F_1$ score) with a top score of $F_1=0.834906$ (1-shot detection). According to the \emph{Skoltech Anomaly Benchmark} leader board\cite{skab}, this is  higher than the previous state-of-the-art (with score of $F_1=0.79$), which is subsequently only trained on normal data (or equivalently 0\% anomalies under our training scheme) with $5$ additional sets of training.

This evidence demonstrates the robustness of our lossy (RDO) approach, which in the case of time-series anomaly detection can deliver state-of-the-art results in a completely \emph{unsupervised} training setting.

\begin{table*}
	\centering
	\begin{tabular}{lccccc}
		\toprule
		Model Type &  Max $F_1$ Score &  Anomaly \% &  $\lambda_1$ &  $\lambda_1$ &  Channel Width \\
		\midrule
		Lossy (RDO) &            0.827350 &        0 &       $1.0\times 10^5$    &   $1.0\times 10^5$    &            128 \\
		Autoencoder (AE) &      0.790485 &          0 &  $0.0\times 10^0$    &   $0.0\times 10^0$    &             30 \\
		Autoencoder (AE) &      0.669088 &          0 &  $0.0\times 10^0$    &   $0.0\times 10^0$    &            128 \\
		Lossy (RDO) &      0.819701 &          1 &        $1.0\times 10^5$   &   $1.0\times 10^5$    &            128 \\
		Autoencoder (AE) &      0.801353 &          1 &  $0.0\times 10^0$    &         $0.0\times 10^0$ &             30 \\
		Autoencoder (AE) &      0.792469 &          1 &             $0.0\times 10^0$ &         $0.0\times 10^0$ &            128 \\
		\textbf{Lossy (RDO)} &      \textbf{0.834906} &          \textbf{5} &        $\mathbf{1.0\times 10^5}$ &    $\mathbf{1.0\times 10^5}$ &            \textbf{128} \\
		Autoencoder (AE) &      0.753554 &          5 &             $0.0\times 10^0$ &         $0.0\times 10^0$ &             30 \\
		Autoencoder (AE) &      0.718331 &          5 &             $0.0\times 10^0$ &         $0.0\times 10^0$ &            128 \\
		Lossy (RDO) &      0.815844 &         10 &        $1.0\times 10^5$ &    $1.0\times 10^5$ &            128 \\
		Autoencoder (AE) &      0.774506 &         10 &             $0.0\times 10^0$ &         $0.0\times 10^0$ &             30 \\
		Lossy (RDO) &      0.776551 &         15 &        $1.0\times 10^5$ &    $1.5\times 10^5$ &            128 \\
		Autoencoder (AE) &      0.742506 &         15 &             $0.0\times 10^0$ &         $0.0\times 10^0$ &            128 \\
		Lossy (RDO) &      0.708883 &         20 &        $1.0\times 10^5$ &    $1.0\times 10^5$ &            128 \\
		Autoencoder (AE) &      0.684747 &         20 &             $0.0\times 10^0$ &         $0.0\times 10^0$ &            128 \\
		Autoencoder (AE) &      0.654526 &         20 &             $0.0\times 10^0$ &         $0.0\times 10^0$ &             30 \\
		Lossy (RDO) &      0.632292 &         25 &        $1.0\times 10^5$ &    $2.0\times 10^5$ &            128 \\
		Autoencoder (AE) &      0.628523 &         25 &             $0.0\times 10^0$ &         $0.0\times 10^0$ &            128 \\
		Autoencoder (AE) &      0.619797 &         25 &             $0.0\times 10^0$ &         $0.0\times 10^0$ &             30 \\
		\bottomrule
	\end{tabular}
	\caption{Experimental Results and their subsequent configurations, best performing model is highlighted}
	\label{tab:results}
\end{table*}

\subsection{Real-time Anomaly detection}\label{sec:multi-shot}
\begin{figure}
	\centering
	\includegraphics[width=1.0\linewidth]{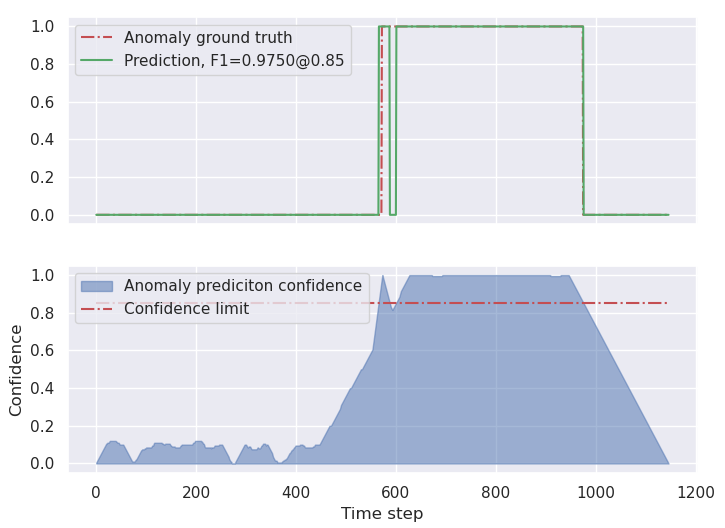}
	\caption{Real time (multi-shot) prediction example, with detection criteria (green solid line \cref{eq:detection_criteria}) based on the confidence score (blue shaded region \cref{eq:confidence_score}) exceeding the limit of 0.85}
	\label{fig:reatime-prediction}
\end{figure}

We can further improve the performance of our RDO anomaly detection scheme by considering the same model used in the 1-shot detection paradigm but applied over a sliding window in \emph{real-time}. Under this (multi-shot) detection setting,  each sample of the signal and its detection score are used to calculate a cumulative ``confidence score'' as seen in \Cref{fig:reatime-prediction}.

For a time $t$, a window $0 < t \leq 200$ is used for anomaly detection and a vector of predictions $d(\mathbf{x})$  is calculated from \cref{eq:1-shot}, we will consider this the 1-shot score and denote further the decomposition $d_t(x_k)$ under the observation that one prediction $d(x_k)$ is valid for the time interval $10k < t \leq 10(k+1)$ such that $d(x_k) = \left[d_{10k}(x_k), \ldots, d_{10(k+1)}(x_k)\right]$, we will then denote the compact notation $d(\mathbf{x}) = \left[d(x_0), \ldots,d(x_{19})\right]$ as $d_{200}(\mathbf{x})$ as the 1-shot score over the interval $0 < t \leq 200$ where
\begin{align}
	d_{200}(\mathbf{x}) = \left[d_{1}(x_0), d_{2}(x_0), \ldots, d_{199}(x_{19}), d_{200}(x_{19}) \right]
\end{align}
For each consecutive time step $t+1$ the prediction windows shifts by 1 such that $d_{201}(\mathbf{x})$ is a prediction for all times $t$ in the interval $1 < t \leq 201$. We can easily see that (baring the initial $t=1$) that for each time step $t$ there will be multiple 1-shot predictions up to a maximum $T=200$, the confidence score (CS) will be the sum of all scores recorded for the instance $t$ such that
\begin{equation}\label{eq:confidence_score}
	CS_t = \kappa\sum^{19}_{0} d_t(x_k)
\end{equation}
where $\kappa = 1/200 \; \forall t > 199, \text{else } \kappa = 1/t $. The the multi-shot anomaly prediction of $H_1$ becomes
\begin{equation}\label{eq:detection_criteria}
	\zeta_t(x) = \begin{cases}
		1       & \quad \text{if } \text{CS}_t > 0.85 \\
		0  & \quad \text{if } \text{CS}_t \leq 0.85
	\end{cases}
\end{equation}
where $\zeta_t(x)$ is the anomaly prediction at time $t$. We can see in \Cref{fig:reatime-prediction}, with even a fairly strict confidence threshold of 0.85, that the score improves to $F_1 = 0.9750$ in this multi-shot scenario compared to an original score $F_1 = 0.834906$ in the 1-shot scenario, using the same architecture and parameterisation! 

\section{Analyses of the Results}\label{sec:analysis}
Our experiments show that the entropy bottleneck in the implementation of our RDO scheme effectively regularises the latent representation in a way that make the whole system robust to the presence of outliers in the training process. This was clearly observed when compared with the expressive AE encoder-decoder network (with 128 channels) that was trained in similar conditions (including anomalies): the AE network fits all data equally showing a sub-optimal anomaly detector performamce.  In contrast,  our RDO scheme does not show this behaviour and, furthermore, it outperforms the-state-of-that-art in anomaly detection with up to 10\% anomalous data. 

The tendency for over-trained autoencoder (AE) models to reconstruct anomalous data was alluded to in \Cref{sec:intro}. It is particularly evident when comparing the results of the autoencoder model (AE) in \Cref{tab:results} with a channel width of 30 \vs 128. The model with the smaller capacity (30 channel) consistently outperforms the larger model in the anomaly detection task. We conjecture that this is due to restriction of the expressiveness of the smaller model, which forces the model to represent exclusively the more 
likely behaviour (i.e the normal patterns). The more expressive architecture trained on the same data with the entropy bottleneck (\ie with compression) has the capacity to robustly learn normality and is effectively regularised by the rate without adversely affecting its reconstruction performance. This contrast in highlighted as the "Anomaly \%" in \Cref{fig:f1scores} increases to 5\% the lossy (RDO) model improves its performance seemingly due to the increase training size and its capacity to robustly filter what is anomalous from that additional data compared to both AE model performances, which degrade sharply. The improved performance of the lossy model, we conjecture, is likely due to the improvement in the signal information with a subsequent smaller increase in the noise, the exploration of the relationship to the lossy model performance and the signal-noise ratio of the source will the subject of future work.

\begin{figure}
	\centering
	\includegraphics[width=1.0\linewidth]{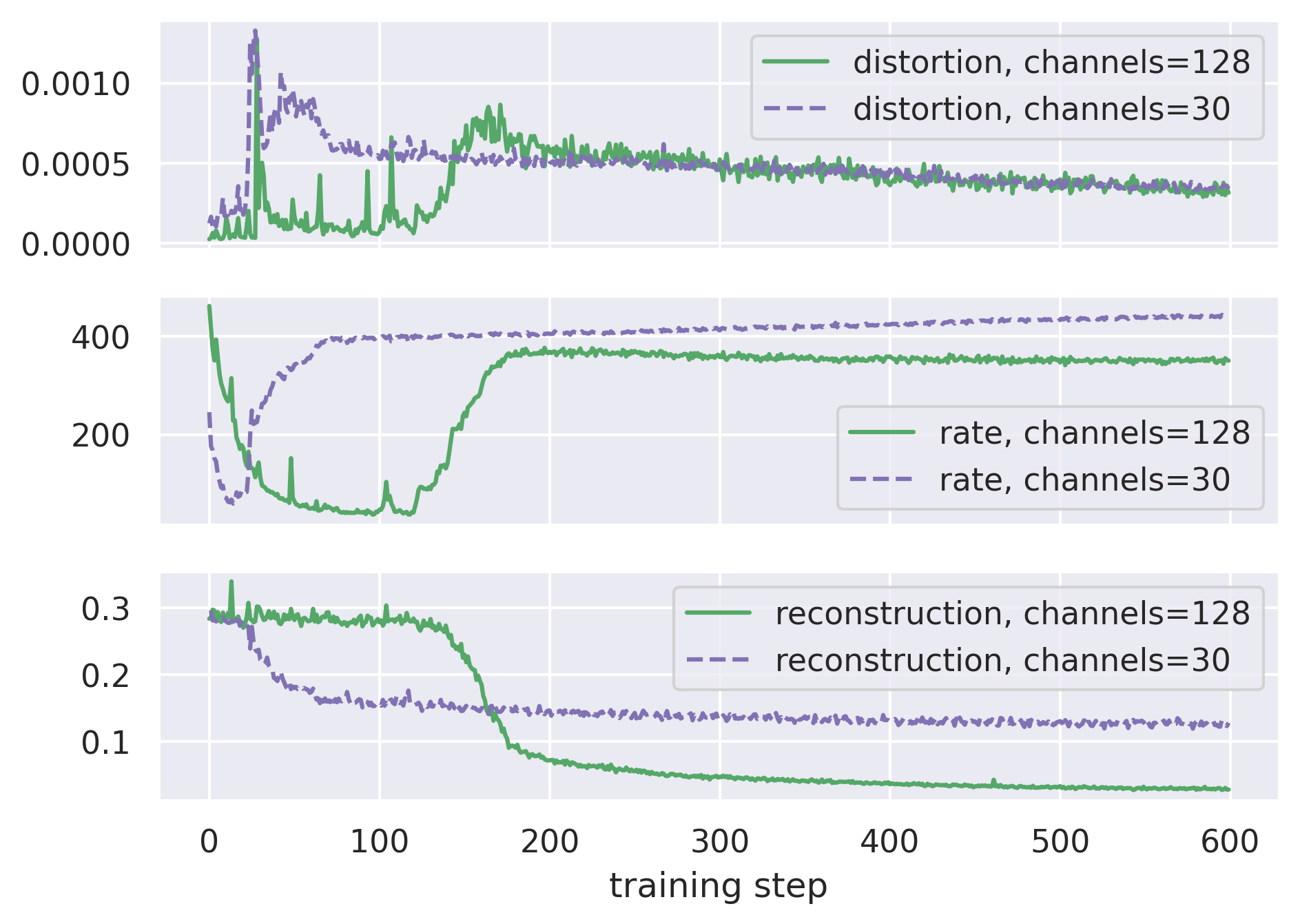}
	\caption{Comparing the 3 loss components unscaled (rate, distortion and reconstruction) of \Cref{eq:rdo_loss} of the best performing model with an identical model but, with a reduced channel capacity: 128 \vs 30. Each model is trained under identical scenarios with 5\% anomalous data corrupting the training data. The model with channel capacity of 128 acheived a $F_1 = 0.834906$, where as the model with the reduced channels only achieved a $F_1=0.642471$ }
	\label{fig:trainingscores}
\end{figure}

An additional benefit to \emph{rate-distortion optimised} (RDO) model, is that during training (using \cref{eq:rdo_loss}), we have access to an efficient way to estimate the rate of our model as it learns an optimal representation.  From this estimation we observed evidence for the effect of the encoder-decoders capacity to retain critical information of the signal. We can see this effect when we compare the three loss metrics (distortion, rate and reconstruction) of \Cref{fig:trainingscores} of two identical models, described in \Cref{sec:TCN}, but with differing channels per layer. The highly expressive model having 128 channels per dilated convolutional layer \vs 30 channels in the less expressive model. We see in \Cref{fig:trainingscores} that, ceteris paribus, the more expressive network (128 channels per layer) converges to a significantly lower estimated rate $R$ while simultaneously achieving a lower reconstruction loss. The requirement for less information to be encoded in the latent variable while simultaneously achieving a better reconstruction seems to imply that information regarding the signal is encoded in the weights of the model themselves.  We conjecture this has a connection to what Ball{\'e} \cite{balle2018variational} alludes to  as ``side information''.

Finally, the notion of ``normality'' is worth discussing in the context of the results demonstrated in \Cref{fig:f1scores} and \Cref{tab:results}. We hypothesise that a lossy model as described in \Cref{sec:lossy} can achieve a robust representation of the underlying signal while filtering unlikely events (such as anomalies), the results we achieve seem to support our hypothesis. One important point to highlight is the degradation of the detection performance of the lossy model as the percentage of anomalies approach 25\%, ultimately converging with the other AE models. The success of the lossy model in robustly detecting anomalies we assert is due to the rate constraint and the restriction of the model to express unlikely events, but events that make up to 1/4th of the data can hardly be called unlikely (and questionably abnormal), for this reason as anomalous data begins to dominate the training set we believe it is natural that the lossy model will begin represent anomalous data in order to improve the average reconstruction in \cref{eq:rdo_loss}. We conjecture that this matter is not of significant practical importance as typically, as highlighted in \cite{Kingma2014}, anomalous events are rare in operational data, and rightly so, intuitively it dose not make sense to define an event that happens 1/4 of the time as abnormal (or anomalous).

\section{Final Discussion}
\begin{itemize}
	\item The basic premise of our scheme is the fact the rate-distortion curve obtained as the solution of \cref{eq:rdo_loss} for different $\lambda_1,\lambda_2 \geq 0$ offers a distinctive signature of the compressibility structure of the regular model $\mu_X$. Therefore, if $\mathbf{x}$ deviates from $H_0$, creating a realisation (a time series signal) that is atypical (in the sense of low probability under $\mu_X$),  this should be expressed (in average) as a salient feature in the sample-wise distortion-rate pair obtained for that signal. The role of the optimal encoder-decoder in defining these deviation patterns is crucial. We argue that the more compressible the source (under $H_0$), the better the capacity of our scheme for discriminating the two hypotheses. 
	
	\item Supporting the previous point, it is well known by the source coding community that an encoder-decoder that is optimal for one model $x\sim \mu_x$ ($H_0$) does not perform optimally when it is adopted to compress the information of a different information source $\tilde{x} \sim q_x \neq \mu_x$ ($H_1$) \cite{Cover2006}. This model miss-match produces an overhead in rate (called redundancy) and an overhead in distortion. These discrepancies are very sensitive to a slight deviation from a reference model. The analysis of these discrepancies has been studied systematically in universal source coding \cite{Gersho1992,Boucheron2009,Davisson1973,Gray1990,Kontoyiannis2000,Rissanen1984,Silva2020}. 
	
	\item There is an interesting connection between the framework proposed in this work for anomaly detection and the problem of discarding invariances (redundancies) observed in the data. We argue that the encoder-decoder has the capacity to look at the most critical features observed in $\mathbf{x}$, discarding irrelevant (infrequent) patterns or self-predictable dimensions of the signal. In that regard, the encoder  is a lossy mapping that discards dimensions of the signal (factors) that are irrelevant for the optimal prediction of $x$ in the MSE sense (under $H_0$). Then all the characteristics of $\mathbf{x}$ that are self-predictable (under $H_1$) have the potential to be removed by solving the lossy compression task in \cref{eq:balle_distortion_loss}. Then, the latent variable $y$ (and consequently $\hat{\mathbf{x}}$) can be insensitive (invariant) to the effects of these discarded redundant factors. Overall, we observe some interesting connections between our anomaly detection application and the lossless prediction problem (under some invariant assumptions) recently presented in \cite{Dubois2021}.
\end{itemize}

{\small
	\bibliographystyle{ieee_fullname}
	\bibliography{LossyAE}

\begin{thebibliography}{10}\itemsep=-1pt

\bibitem{Achille2018}
A. Achille and S. Soatto.
\newblock Information dropout: Learning optimal representations through noisy
  computations.
\newblock {\em IEEE Transactions on Pattern Analysis and Machine Intelligence},
  40(12):2897 -- 2905, January 2018.

\bibitem{akcay2018ganomaly}
Samet Akcay, Amir Atapour-Abarghouei, and Toby~P Breckon.
\newblock Ganomaly: Semi-supervised anomaly detection via adversarial training.
\newblock In {\em Asian conference on computer vision}, pages 622--637.
  Springer, 2018.

\bibitem{Alemi2017}
A. Alemi, I. Fischer, J. Dillon, and K. Murphy.
\newblock Deep variational information bottleneck.
\newblock In {\em in Proc. Int. Conf. Learn. Represent. (ICLR),}, pages
  368--377, April 2017.

\bibitem{balle2018variational}
Johannes Ball{\'e}, David Minnen, Saurabh Singh, Sung~Jin Hwang, and Nick
  Johnston.
\newblock Variational image compression with a scale hyperprior.
\newblock {\em International Conference on Learning Representations (ICLR)},
  2018.

\bibitem{baur2018deep}
Christoph Baur, Benedikt Wiestler, Shadi Albarqouni, and Nassir Navab.
\newblock Deep autoencoding models for unsupervised anomaly segmentation in
  brain mr images.
\newblock In {\em International MICCAI brainlesion workshop}, pages 161--169.
  Springer, 2018.

\bibitem{bergmann2018improving}
Paul Bergmann, Sindy L{\"o}we, Michael Fauser, David Sattlegger, and Carsten
  Steger.
\newblock Improving unsupervised defect segmentation by applying structural
  similarity to autoencoders.
\newblock {\em arXiv preprint arXiv:1807.02011}, 2018.

\bibitem{Boucheron2009}
S. Boucheron, A. Garivier, and E. Gassiat.
\newblock Coding on countably infinite alphabets.
\newblock {\em IEEE Transactions on Information Theory}, 55(1):358--373, 2009.

\bibitem{bronstein2021geometric}
Michael~M Bronstein, Joan Bruna, Taco Cohen, and Petar Veli{\v{c}}kovi{\'c}.
\newblock Geometric deep learning: Grids, groups, graphs, geodesics, and
  gauges.
\newblock {\em arXiv preprint arXiv:2104.13478}, 2021.

\bibitem{chen2022utrad}
Liyang Chen, Zhiyuan You, Nian Zhang, Juntong Xi, and Xinyi Le.
\newblock Utrad: Anomaly detection and localization with u-transformer.
\newblock {\em Neural Networks}, 147:53--62, 2022.

\bibitem{chen2018isolating}
Ricky~TQ Chen, Xuechen Li, Roger~B Grosse, and David~K Duvenaud.
\newblock Isolating sources of disentanglement in variational autoencoders.
\newblock {\em Advances in neural information processing systems}, 31, 2018.

\bibitem{Cover2006}
T.~M. Cover and J.~A. Thomas.
\newblock {\em Elements of Information Theory}.
\newblock Wiley Interscience, New York, second edition, 2006.

\bibitem{Davisson1973}
Lee~D. Davisson.
\newblock Universal noiseless coding.
\newblock {\em IEEE Transactions on Information Theory}, IT-19(6):783--795,
  1973.

\bibitem{Dubois2021}
Y. Dubois, B. Bloem-Reddy, K. Ullrich, and C.~J. Maddison.
\newblock Lossy compression for losless prediction.
\newblock In {\em at ICLR 2021 neural compression workshop}, pages 1--26, 2021.

\bibitem{geiger2020tadgan}
Alexander Geiger, Dongyu Liu, Sarah Alnegheimish, Alfredo Cuesta-Infante, and
  Kalyan Veeramachaneni.
\newblock Tadgan: Time series anomaly detection using generative adversarial
  networks.
\newblock In {\em 2020 IEEE International Conference on Big Data (Big Data)},
  pages 33--43. IEEE, 2020.

\bibitem{Gersho1992}
A. Gersho and R.M. Gray.
\newblock {\em Vector Quantization and Signal Compression}.
\newblock Norwell, MA: Kluwer Academic, 1992.

\bibitem{Gray1990}
R.M. Gray.
\newblock {\em Source Coding Theory}.
\newblock Norwell, MA: Kluwer Academic, 1990.

\bibitem{higgins2016beta}
Irina Higgins, Loic Matthey, Arka Pal, Christopher Burgess, Xavier Glorot,
  Matthew Botvinick, Shakir Mohamed, and Alexander Lerchner.
\newblock beta-vae: Learning basic visual concepts with a constrained
  variational framework.
\newblock 2016.

\bibitem{kato2020rate}
Keizo Kato, Jing Zhou, Tomotake Sasaki, and Akira Nakagawa.
\newblock Rate-distortion optimization guided autoencoder for isometric
  embedding in euclidean latent space.
\newblock In {\em International Conference on Machine Learning}, pages
  5166--5176. PMLR, 2020.

\bibitem{skab}
Iurii~D. Katser and Vyacheslav~O. Kozitsin.
\newblock Skoltech anomaly benchmark (skab).
\newblock \url{https://www.kaggle.com/dsv/1693952}, 2020.

\bibitem{kim2018disentangling}
Hyunjik Kim and Andriy Mnih.
\newblock Disentangling by factorising.
\newblock In {\em International Conference on Machine Learning}, pages
  2649--2658. PMLR, 2018.

\bibitem{Kingma2014}
D. Kingma and M. Welling.
\newblock Auto-encoding variational bayes.
\newblock In {\em in Proc. Int. Conf. Learn. Represent. (ICLR),}, 2014.

\bibitem{Kontoyiannis2000}
Ioannis Kontoyiannis.
\newblock Pointwise redundancy in lossy data compression and universal lossy
  data compression.
\newblock {\em IEEE Transactions on Information Theory}, 46(1):136--152,
  January 2000.

\bibitem{le2020learning}
Xinyi Le, Junhui Mei, Haodong Zhang, Boyu Zhou, and Juntong Xi.
\newblock A learning-based approach for surface defect detection using small
  image datasets.
\newblock {\em Neurocomputing}, 408:112--120, 2020.

\bibitem{liu2022time}
Shenghua Liu, Bin Zhou, Quan Ding, Bryan Hooi, Zheng bo Zhang, Huawei Shen, and
  Xueqi Cheng.
\newblock Time series anomaly detection with adversarial reconstruction
  networks.
\newblock {\em IEEE Transactions on Knowledge and Data Engineering}, 2022.

\bibitem{macikag2021unsupervised}
Piotr~S Maci{\k{a}}g, Marzena Kryszkiewicz, Robert Bembenik, Jesus~L Lobo, and
  Javier Del~Ser.
\newblock Unsupervised anomaly detection in stream data with online evolving
  spiking neural networks.
\newblock {\em Neural Networks}, 139:118--139, 2021.

\bibitem{Rissanen1984}
J. Rissanen.
\newblock Universal coding, information, prediction, and estimation.
\newblock {\em IEEE Transactions on Information Theory{\c c}},
  IT-30(4):629--636, July 1984.

\bibitem{rolinek2019variational}
Michal Rolinek, Dominik Zietlow, and Georg Martius.
\newblock Variational autoencoders pursue pca directions (by accident).
\newblock In {\em Proceedings of the IEEE/CVF Conference on Computer Vision and
  Pattern Recognition}, pages 12406--12415, 2019.

\bibitem{salehi2021multiresolution}
Mohammadreza Salehi, Niousha Sadjadi, Soroosh Baselizadeh, Mohammad~H Rohban,
  and Hamid~R Rabiee.
\newblock Multiresolution knowledge distillation for anomaly detection.
\newblock In {\em Proceedings of the IEEE/CVF conference on computer vision and
  pattern recognition}, pages 14902--14912, 2021.

\bibitem{schlegl2019f}
Thomas Schlegl, Philipp Seeb{\"o}ck, Sebastian~M Waldstein, Georg Langs, and
  Ursula Schmidt-Erfurth.
\newblock f-anogan: Fast unsupervised anomaly detection with generative
  adversarial networks.
\newblock {\em Medical image analysis}, 54:30--44, 2019.

\bibitem{schlegl2017unsupervised}
Thomas Schlegl, Philipp Seeb{\"o}ck, Sebastian~M Waldstein, Ursula
  Schmidt-Erfurth, and Georg Langs.
\newblock Unsupervised anomaly detection with generative adversarial networks
  to guide marker discovery.
\newblock In {\em International conference on information processing in medical
  imaging}, pages 146--157. Springer, 2017.

\bibitem{silva2022interplay}
Jorge Silva and Felipe Tobar.
\newblock On the interplay between information loss and operation loss in
  representations for classification.
\newblock In {\em International Conference on Artificial Intelligence and
  Statistics}, pages 4853--4871. PMLR, 2022.

\bibitem{Silva2020}
Jorge~F Silva and Pablo Piantanida.
\newblock Universal weak variable-length source coding on countably infinite
  alphabets.
\newblock {\em IEEE Transactions on Information Theory}, 66(1):649--668,
  January 2020.

\bibitem{solch2016variational}
Maximilian S{\"o}lch, Justin Bayer, Marvin Ludersdorfer, and Patrick van~der
  Smagt.
\newblock Variational inference for on-line anomaly detection in
  high-dimensional time series.
\newblock {\em arXiv preprint arXiv:1602.07109}, 2016.

\bibitem{thill2020time}
Markus Thill, Wolfgang Konen, and Thomas B{\"a}ck.
\newblock Time series encodings with temporal convolutional networks.
\newblock In {\em International Conference on Bioinspired Methods and Their
  Applications}, pages 161--173. Springer, 2020.

\bibitem{Tishby2015}
N. Tishby and N. Zaslavsky.
\newblock Deep learning and the information bottleneck principle.
\newblock In {\em Information Theory Workshop}, pages 1--5, 2015.

\bibitem{oord2016wavenet}
A{\"a}ron Van Den~Oord, Sander Dieleman, Heiga Zen, Karen Simonyan, Oriol
  Vinyals, Alex Graves, Nal Kalchbrenner, Andrew Senior, and Koray Kavukcuoglu.
\newblock Wavenet: A generative model for raw audio.
\newblock {\em arXiv preprint arXiv:1609.03499}, 2016.

\bibitem{van2016conditional}
Aaron Van~den Oord, Nal Kalchbrenner, Lasse Espeholt, Oriol Vinyals, Alex
  Graves, et~al.
\newblock Conditional image generation with pixelcnn decoders.
\newblock {\em Advances in neural information processing systems}, 29, 2016.

\bibitem{van2016pixel}
A{\"a}ron Van Den~Oord, Nal Kalchbrenner, and Koray Kavukcuoglu.
\newblock Pixel recurrent neural networks.
\newblock In {\em International conference on machine learning}, pages
  1747--1756. PMLR, 2016.

\bibitem{yu2015multi}
Fisher Yu and Vladlen Koltun.
\newblock Multi-scale context aggregation by dilated convolutions.
\newblock {\em arXiv preprint arXiv:1511.07122}, 2015.

\bibitem{zhang2021soft}
Haodong Zhang, Yongquan Chen, Bin Liu, Xinping Guan, and Xinyi Le.
\newblock Soft matching network with application to defect inspection.
\newblock {\em Knowledge-Based Systems}, 225:107045, 2021.

\end{thebibliography}
}

\end{document}